\documentclass[10pt,conference,a4paper]{IEEEtran}
\ifCLASSINFOpdf
  \usepackage[pdftex]{graphicx}
\else
\fi
\usepackage{caption}
\usepackage{subcaption}
\usepackage[cmex10]{amsmath}
\usepackage{amssymb}

\newcommand\norm[1]{\left\lVert#1\right\rVert}
\newcommand{\R}{\mathbb{R}}
\usepackage{mathtools}
\usepackage{algorithm}
\usepackage{algorithmic}
\usepackage{array}
\usepackage{cite}

\begin{document}

\title{A Novel Actor Dual-Critic Model for Remote Sensing Image Captioning}

\author{\IEEEauthorblockN{Ruchika Chavhan \IEEEauthorrefmark{1},
Biplab~Banerjee\IEEEauthorrefmark{1},
Xiao Xiang Zhu\IEEEauthorrefmark{2}, 
Subhasis~Chaudhuri\IEEEauthorrefmark{1}}
\IEEEauthorblockA{\IEEEauthorrefmark{1} Indian Institute of Technology Bombay, India\\}
\IEEEauthorblockA{\IEEEauthorrefmark{2}Signal Processing in Earth Observation, Technical University of Munich, Germany\\}}

\maketitle

\begin{abstract}
We deal with the problem of generating textual captions from optical remote sensing (RS) images using the notion of deep reinforcement learning. Due to the high inter-class similarity in reference sentences describing remote sensing data, jointly encoding the sentences and images encourages prediction of captions that are semantically more precise than the ground truth in many cases. To this end, we introduce an Actor Dual-Critic training strategy where a second critic model is deployed in the form of an encoder-decoder RNN to encode the latent information corresponding to the original and generated captions. While all actor-critic methods use an actor to predict sentences for an image and a critic to provide rewards, our proposed encoder-decoder RNN guarantees high-level comprehension of images by sentence-to-image translation. We observe that the proposed model generates sentences on the test data highly similar to the ground truth and is successful in generating even better captions in many critical cases. Extensive experiments on the benchmark Remote Sensing Image Captioning Dataset (RSICD) and the UCM-captions dataset confirm the superiority of the proposed approach in comparison to the previous state-of-the-art where we obtain a gain of sharp increments in both the ROUGE-L and CIDEr measures.
\end{abstract}

\IEEEpeerreviewmaketitle

\section{Introduction}

Image Captioning is the task of generating a natural language description for a given image. Describing the images in properly framed sentences is a task that humans can perform with utmost ease \cite{Feifei2007WhatDW}, but was an inconceivable task for computers before the advent of deep learning. The primary requirements to automatically generate a caption are capturing the essence of the image and describing the correlations between objects to localize them in the image. 
Image captioning is more difficult than other tasks  \cite{ILSVRC15} that the computer vision community has concentrated on, as it involves the integration of semantic interpretation of an image into textual explanations.

Image captioning data is derived from different modalities: images that are represented by pixel intensities and textual descriptions represented as discrete word count vectors. The center of image captioning is recognizing the collective interpretation of these various modalities. The problem setting requires studying an abstract interpretation of the contents of the image and the semantic associations between the objects that can be constructed as a natural language description. Therefore, the problem of image captioning was defined as: Given an image $I$, a model is trained to maximize the likelihood $p(W|I)$ where $W = {w_1, w_2, w_3, ... w_n}$ where all $w_i$ are words from a pre-defined vocabulary. One of the first methods to describe images in human interpretable language  \cite{7298935} employed a Convolutional Neural Network (CNN) as an encoder to extract meaningful features from an image in the form of a fixed vector and a recurrent neural network (RNN) as a decoder to generate sentences. The primary inspiration of this model arises from the intuition that images can be translated to sentences, similar to the problem of machine translation \cite{conf/emnlp/ChoMGBBSB14} which follows an encoder-decoder architecture to translate the source sentence into target format through a bottleneck latent space. The goal of subsequent works in image captioning \cite{7298932, 10.5555/3045118.3045336} has been to produce more diverse captions that also accurately represent the image content. The aim of most image captioning works is to produce multiple contextual explanations that novelly manifest the local associations of objects in the image.

Reinforcement Learning \cite{10.5555/3312046} is a domain of Machine Learning that enables an agent to explore an environment by performing an action determined by a policy. While Deep Learning offers the best set of models for learning representations of multi-modal data, Reinforcement Learning is a framework for learning sequential decision-making tasks. Therefore, RL is the  mainstream algorithm used to solve complex environments in different games to achieve super-human performance\cite{10.1145/3206157.3206174}, robotics\cite{10.5555/2946645.2946684}, and recommendation systems. Most RL agents acquire a stochastic mapping of states and actions, called the policy and pursue a trajectory by carrying out actions determined by this policy to maximize the total expected return in a given time phase. A model-free RL agent does not pre-specify a structural model of the environment, instead, it gradually learns the best policies based on trial and error and adequate exploration of the environment \cite{10.5555/1620270.1620297}. Most environments consist of multi-modal data and a single joint representation is learnt by an RL agent as a decision-making framework.

All reinforcement learning-based approaches are an exploration-based approach in which the agent first collects knowledge about the entire environment and state-space to anticipate reward-maximizing behavior. The agent is able to make highly optimized decisions in that way. Therefore, addressing the task of image captioning in the Reinforcement learning domain enables the generation of semantically more coherent captions. All supervised learning methods aim to generate sentences that are exactly identical to ground truth, while policies trained in an RL setup allow one to predict sentences much better than ground truth. Policy-generated natural language descriptions can be dramatically enhanced by increasing the degree of environmental analysis consisting of paired images and captions.
 
Recently, acquisition of an unprecedented volume of satellite images from different sensors is observed.  The major product derived from the satellite images has been the land-cover maps which assigns semantic class-labels at the pixel locations. However, the task of pixel-labeling is redundant for some emergency applications where an overall description of the scene under consideration is encouraged. Under this premise, the task of generating automatic captions for satellite scenes holds much potential. However, there does not exist extensive literature in this front specifically for remote sensing data.

In this paper, we implement an Actor Dual-Critical (ADC) training setup to address the issue of high inter-class image and caption similarity in satellite data. We have performed our proposed experiment on the Remote Sensing Image Captioning Dataset (RSICD) \cite{lu2017exploring} and the UCM-captions dataset \cite{7546397, inproceedings}. The key problem in the datasets for many images is the strong inter-class similarity and the identical reference sentences.
The contributions and results of the paper are summarized as follows:
\begin{itemize}
\setlength\itemsep{1pt}
\item To the best of our knowledge, our approach is the first to use Reinforcement Learning to produce captions on remote sensing images. Unlike all existing methods, we employ an additional encoder-decoder RNN as a critic for Actor Dual-Critic (ADC) training setup.  This critic plays a key role in creating a variety of different sentences that represent identical visual perception.
\item The captions predicted by our proposed model are more semantically related to the objects in the image, explicitly describe object localization, and specifically focus on the existence of the entities in the image.
\item We also perform cross-dataset captioning and obtain superior results on the RSICD dataset by models trained on the UCM-captions data set to demonstrate the generalization capability of a policy trained using the ADC setup as compared to previous state-of-the-art methods.
\end{itemize}

\section{Related Work}

Actor-critic methods in the Reinforcement Learning aim to train the participant in choosing actions taking into account the critic's reward.
In the context of image captioning implemented in an actor-critic training strategy \cite{ZhangSLXGYH17}, given an image $I$ and partially generated sentence $S = (w_{1},w_{2}, ... , w_{t})$ , $w_{t+1}$ is viewed as an action that the policy predicts. 
An actor's job is to learn a policy $\pi(a_{t} | s_{t-1})$, where $a_{t}$ is the action performed and $s_{t}$ is the state at time $t$. The critic provides the value function $v_{\theta}^{\pi}$, where $\theta$ denotes the parameters of the value network. The reward metric for the generated caption is the score evaluated by ROUGE-L and $v_{\theta}^{\pi}$ is used as an advantage baseline for the advantage factor $A^{\pi}(s_{t}|a_{t+1}) = (\gamma^{T-t-1}r_{T} - v_{\theta}^{\pi})$. Subsequently, the actor is trained to optimize expected value of reward over the trajectory using a cross-entropy loss and the advantage factor using the REINFORCE Algorithm where gradient are given by $ \mathbf{E}[\sum_{t=0}^{T} A^{\pi}(s_{t}|a_{t+1}) \nabla \log\pi(a_{t}| s_{t-1})]$. Hence, this training method is named Advantage-Actor Critic (A2C). Instead of calculating a value function for baseline, the test-time reward metric can also be used a baseline to normalize the rewards experienced during training, thus eliminating the need for a critic, making the training setup self-critical \cite{RennieMMRG16, abs-1904-06861}.

Recent work on remote sensing image captioning \cite{7546397, 8900503} involves using deep multi-modal networks and analyzing the quality of generated captions by experimenting with various types of CNN, RNN, and LSTM combinations. Similar experiments were performed on the RSICD dataset \cite{lu2017exploring} to produce accurate captions. Because of the wide area of the Earth's surface covered by remote sensing images, the main criteria for caption generation are to recognize semantic uncertainty in remote sensing images by analyzing key instances \cite{7891049} and performing image context and landscape analysis.


\begin{figure*}[t]
\begin{subfigure}{.47\textwidth}
\centering
  \includegraphics[width=.8\linewidth, height=2.0cm]{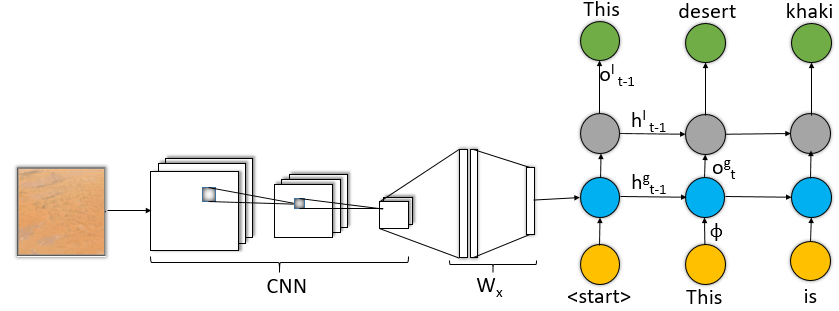}
  \caption{Working of the actor. The words are converted into the embedding space (not shown) before being fed into the GRU (denoted by blue) and LSTM (denoted by grey). }
  \label{fig:actor}
\end{subfigure}
\hspace{20pt}
\begin{subfigure}{.47\textwidth}
\centering
  \includegraphics[width=.8\linewidth, height=2.0cm]{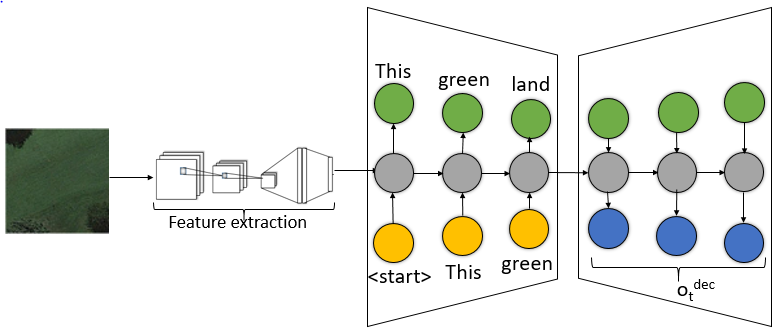}
  \caption{Working of the proposed Encoder-Decoder RNN critic. The outputs and the hidden states of the encoder RNN are transformed by $\psi_{1}$ and $\psi_{2}$ (not shown) and serve as input to the decoder RNN.  }
  \label{fig:critic}
\end{subfigure}
\caption{Working Principle of the proposed ADC setup}
\end{figure*}

\section{Proposed Methodology} \label{sec:sec_method}
We first discuss the working principles of our proposed Actor Dual-Critic (ADC) training and elaborate on the three components employed in the process. 
We formulate the problem statement as follows: Given an image $I$, the model is expected to generate a natural language description $S = (w_{1}, w_{2}, ... , w_{n})$ where $w_{i}$ is a word from a pre-defined vocabulary. The actor generates a sentence given the image and the critics provide rewards based on the quality and relevance of this sentence. We utilize two critics: an RNN critic and an encoder-decoder RNN critic to provide two rewards to update parameters of the actor using the REINFORCE Algorithm. The training algorithm is mentioned in Algorithm \ref{alg:the_alg}.

\vspace{3pt}
\noindent
\textbf{Learning the Policy:} We propose a three-component model for the actor to learn a policy. The actor provides a measure of confidence $q_{\pi}(a_{t}|s_{t})$ to predict the next action $\mathbf{a_{t}} = \mathbf{w_{t+1}} \in \R^{d}$ according to the current state. For the extraction of features from an image, we utilize a pre-trained CNN. The extracted features $\mathbf{f} \in \R^{n}$ are passed as an input to a Gated Recurrent Unit (GRU). The hidden state of GRU  $\mathbf{h_{t}^{g}} \in \R^{n}$ evolves with time as the predicted words $w_{t-1}$ are fed into it. The GRU acts indirectly as the generator of a context vector \cite{DBLP:conf/iclr/2015} to assist the decoder to surmount difficulties in learning due longer sentences. We observed a significant increase in performance by replacing the normal LSTM with a Layer Norm LSTM \cite{Ba2016LayerN} with dropout. The following equations explain the functionality of the actor:

\begin{equation}
    \begin{aligned}
    &f = W_{x}(\text{CNN}(I))  \\
    &\phi_{0} = f\\
    &o_{t}^{g}, h_{t}^{g} = \text{GRU}(\phi_{t-1}, h_{t-1}^{g})\\
    &o_{t}^{l}, h_{t}^{l} = \text{LSTM}(o_{t}^{g}, h_{t-1}^{l})\\
    &q_{\pi}(a_{t}|s_{t}) = \psi({o_{t}^{l}})\\
    &\phi_{t} = \zeta(w_{t-1})
    \end{aligned}
\end{equation}

Here, $W_{x}$ is the weight of the linear embedding model
of the CNN, $ o_{t}^{g}$ and $ o_{t}^{l}$ are the outputs of the GRU and LSTM respectively at time step $t$. $\mathbf{\psi} \colon \R^{n} \mapsto \R^{d}$ is a non-linear function that transforms the output of the LSTM to a space whose dimension is equal to the dimension of the vocabulary. $\mathbf{\zeta} \colon \R^{d} \mapsto \R^{n}$ denotes the embedding model to represent words in a common embedding space.  We denote the policy network by $\pi(a_{t}| s_{t-1})$. Please refer to Figure \ref{fig:actor} for complete description and pipeline of the model.

The model is trained to optimize the objective:
\vspace{-5pt}
\begin{equation} \label{actor_loss}
   \min_{\pi}\sum_{t=0}^{T} \log(q_{\pi}(a_{t}|s_{t}))
\end{equation}
\textbf{Value Network:} This critic consists of an RNN which outputs a value function $v_{\theta}^{\pi}(s_{t})$ given the words predicted by the current policy and features extracted by the CNN in Figure \ref{fig:actor}. We initialize the hidden state of the RNN by these extracted features. We will denote this critic as $V(s_{t})$. The output of this network is directed to be the expected value of future rewards $ \mathbb{E}[\sum_{l=0}^{T-t-1} \gamma^{l} r_{t+l+1} | a_{t+1}, ... , a_{T} \sim \pi, I]$ for choosing a state $s_{t}$ given the current policy $\pi$. We set $\gamma = 1$ similar to \cite{ZhangSLXGYH17} and calculate the reward $r_{T}$ after the prediction of the entire sequence implying $r_{t} = 0$, $\forall$ $t < T$.

The reward $r_{T}$ for the entire generated sentence is obtained by using the evaluation scores of ROUGE-L or BLEU. We observed more stable training using the Huber Loss instead of the norm of the difference between the reward and $v_{\theta}$: 

\begin{equation} 
\label{eqn_rnn_loss}
L = 
    \begin{dcases}
        ||v_{\theta}^{\pi} - r_{T}||^{2} & ||v_{\theta}^{\pi} - r_{T}||\leq \delta = 0.5\\
         \delta ||v_{\theta}^{\pi} - r_{T}|| - \frac{1}{2}\delta^{2} &
         \text{otherwise}  \\
    \end{dcases}
\end{equation}

\noindent
{\textbf{Encoder-Decoder LSTM Critic:}} The addition of this critic is the main contribution of this paper. The intuition behind this critic is as follows:
\begin{itemize}
    \item Theoretically, image captioning is defined as translation of images into sentences that aptly describe the images. We hypothesize that the sentences translated back into images should generate a quantity which closely resembles the features extracted from images. 
    \item This procedure ensures maximum accumulation of information about the image through a textual description relatively similar to the information captured by ground truth sentences. 
\end{itemize} 
Please refer to Figure \ref{fig:critic} for the entire pipeline of this model. The working principle of this critic denoted by $D(S)$ is governed by the following equations:
\begin{equation} \label{equations_enc}
    \begin{aligned}
    &h_{0}^{enc} = W_{x}(\text{CNN}(I)) \\
    &\eta_{t} = \zeta(S)\\
    &o_{t}^{enc}, h_{t}^{enc} = \text{RNN}_{enc}(\eta_{t}, h_{t-1}^{enc})\\
    &h_{0}^{dec} = \psi_{2}(h_{T}^{enc})\\
    &i_{1}^{dec} = \psi_{1}(o_{T}^{enc})\\
    &o_{t}^{dec}, h_{t}^{dec} = \text{RNN}_{dec}(i_{t}^{enc}, h_{t-1}^{dec})
    \end{aligned}
\end{equation}

Here, RNN$_{enc}$ and RNN$_{dec}$ are the encoder-decoder RNN respectively. $S = (w_{1}, w_{2}, ... , w_{T})$ denotes a natural language description of the image. We have used $\psi_{1}, \psi_{2} \colon \R^{n} \mapsto \R^{n}$ as a linear function with dimension equal to that of the embedding space of sentences, along with the ReLU activation function.

This critic is trained by optimizing the mean squared error (MSE) between the output of the decoder and the features:

\begin{equation} \label{loss_endec}
    L = (\frac{\sum_{t=0}^{T}o_{t}^{dec}}{|S|} - f)^{2}
\end{equation}
Accuracy for the decoder output is given by the cosine similarity between the output of the decoder and features:
\begin{equation}
  A_{gen} = \frac{\frac{\sum_{t=0}^{T}o_{t}^{dec}}{|S|} f}{\norm{f}\norm{\frac{\sum_{t=0}^{T}o_{t}^{dec}}{|S|}}}
\end{equation}
 $A_{gen}$ and $A_{orig}$ are the accuracies of the network when captions generated by the actor and ground truth captions are fed into the encoder respectively.
We defined an advantage factor for this critic to be:
\begin{equation} \label{eq_ad2}
    A_{ed} = A_{gen} - \delta_{t} A_{orig}
\end{equation}

The encoder-decoder critic is pre-trained on features extracted and corresponding original sentences to learn a ground truth latent representation.  
We observed that $\delta_{t} = 1$ initially leads to exploding gradients and a non-converging policy. Therefore, we begin with $\delta_{t} = 0.01$ and increase it linearly to 1.0 over the epochs. For generating captions for a test image, we pass the image $I_{test}$ to the actor to generate sentences till the $<$end$>$ token is encountered. 
\section{Experiments and Results}
\label{sec:sec_result}

In this section, we study the performance of our model and it's results on two remote sensing image captioning datasets: RSICD and UCM-captions. We also visualise the output of our novel critic and analyze the validity of it's working principle. 

\begin{algorithm}[H]
 \caption{Training Algorithm}
 \begin{algorithmic}[1]
 \label{alg:the_alg}
 \renewcommand{\algorithmicrequire}{\textbf{Input:}}
 \REQUIRE Pre-trained models $\pi(a_{t}|s_{t-1})$, $V(s_{t})$ and $D(S)$ using the objectives given by the equations \ref{actor_loss}, \ref{eqn_rnn_loss} and \ref{loss_endec} respectively as done in \cite{ZhangSLXGYH17}.
  \FOR {$episode = 1$ to total episodes}
  \STATE Given an Image $I$ sample action $(a_{1}, a_{2} ,..., a_{T})$ from the current policy using a multinomial distribution given by $q_{\pi}(s_{t}|a_{t})$;
  \STATE Calculate advantage factor $A^{\pi}$ using the reward $r_{T}$ for the value network;
  \STATE  Update the parameters of the policy using $A^{\pi}$ by the REINFORCE Algorithm;
  \STATE Update parameters of the critic by optimising Eq. \ref{eqn_rnn_loss};
  \STATE Calculate advantage factor $A_{ed}$ using the encoder-decoder critic;
 \STATE Update the parameters of the policy using $A_{ed}$ by the REINFORCE Algorithm;
 \STATE Update parameters of the critic using $A_{orig}. $
  \ENDFOR
 \end{algorithmic}
 \end{algorithm}

\begin{table*}[t] 
\caption{Results of ADC setup on the RSICD dataset}
\begin{tabular}{|p{2.2cm}||p{1.5cm}|p{1.5cm}|p{1.5cm}|p{1.5cm}|p{2.0cm}|p{2.0cm}|p{2.0cm}| }
\hline
Metric & B-1 & B-2 & B-3 & B-4 & METEOR \cite{10.5555/1626355.1626389} & ROUGE-L & CIDEr\cite{Vedantam2014CIDErCI}\\
\hline
MM \cite{lu2017exploring}  & 0.57905 & 0.41871 & 0.32628 & 0.26552 & 0.26103 & 0.51913 & 2.05261 \\
\hline
SA \cite{lu2017exploring} & 0.65638 & 0.51489 & 0.41764 & 0.34464 & 0.32924 & 0.61039 &1.87415 \\
\hline
HA \cite{lu2017exploring}& 0.68968 & 0.5446 & 0.44396 & 0.36895 & \textbf{0.33521} & 0.62673 & 1.98312  \\
\hline
A2C \cite{ZhangSLXGYH17} & 0.60157 & 0.41991 &0.364516 & 0.28788 & 0.19382 & 0.63185 & 2.098 \\
\hline
Ours & \textbf{0.73973} & \textbf{0.55259} & \textbf{0.46353} & \textbf{0.41016} & 0.22126 & \textbf{0.71311} & \textbf{2.243}  \\
\hline
\end{tabular}
\label{Tab:table_res}
\end{table*}

\begin{table*}[t] 
\caption{Results of ADC setup on the UCM dataset}
\begin{tabular}{|p{2.2cm}||p{1.5cm}|p{1.5cm}|p{1.5cm}|p{1.5cm}|p{2.0cm}|p{2.0cm}|p{2.0cm}| }
\hline
Metric & B-1 & B-2 & B-3 & B-4 & METEOR \cite{10.5555/1626355.1626389} & ROUGE-L & CIDEr\cite{Vedantam2014CIDErCI}\\
\hline
MM \cite{lu2017exploring}  &  0.37066 & 0.32344 & 0.32346
 & 0.23259 & 0.40476 & 0.4236 & 1.708
 \\
\hline
SA \cite{lu2017exploring} & 0.79693 & 0.71345 & 0.6514 & 0.59895 & 0.74952 & 0.41676 & 2.12846
 \\
\hline
HA \cite{lu2017exploring}&  0.78498 & 0.70929 & 0.65182 & 0.60167 & 0.77357 & 0.43058 & 2.19594
  \\
\hline
A2C \cite{ZhangSLXGYH17} &  0.373089 & 0.23776 & 0.15857 & 0.12222 & 0.39645 & 0.35989 & 2.381
 \\
\hline
Ours &  \textbf{0.85330} & \textbf{0.75679} & \textbf{0.67854} & \textbf{0.61165} & \textbf{0.83242} & \textbf{0.80872} & \textbf{4.865}
 \\
\hline
\end{tabular}
\label{Tab:tableUCM}
\end{table*}

\begin{table*}[t] 
\caption{Results of cross dataset captioning on the RSICD dataset }
\begin{tabular}{|p{2.2cm}||p{1.5cm}|p{1.5cm}|p{1.5cm}|p{1.5cm}|p{2.0cm}|p{2.0cm}|p{2.0cm}| }
\hline
Metric & B-1 & B-2 & B-3 & B-4 & METEOR \cite{10.5555/1626355.1626389} & ROUGE-L & CIDEr\cite{Vedantam2014CIDErCI}\\
\hline
MM \cite{lu2017exploring}  & 0.19618 & 0.01481 & 0.00721
 & 0.00445 & 0.07416 & 0.2457 & 0.08015
 \\
\hline
A2C \cite{ZhangSLXGYH17} &  0.19405 & 0.04137 & 0.00714 & 0.00175 & 0.18855 & 0.1846 & 0.961
 \\
\hline
Ours &  \textbf{0.38810} & \textbf{0.08643} & \textbf{0.019065} & \textbf{0.00608} & \textbf{0.23964} & \textbf{0.2888} & \textbf{2.013}
 \\
\hline
\end{tabular}
\label{Tab:tableCross}
\end{table*}

\begin{figure*}[h]
\begin{subfigure}{.47\textwidth}
  \centering
  \includegraphics[width=\linewidth]{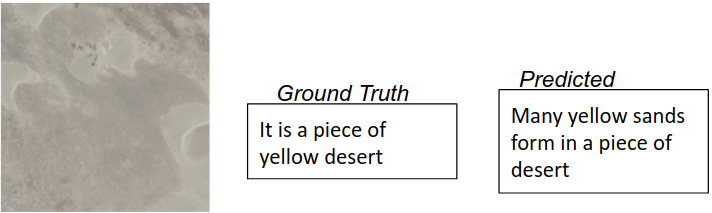}
  \caption{Original Image with ground truth and predicted sentences. The image is from the class 'desert'. A test image from the same class but with the same captions as the test input image are passed to the encoder for this experiment. The test image used for this case belongs to the class 'desert'. }
  \label{fig:example_image}
\end{subfigure}
\hspace{30pt}
\begin{subfigure}{.47\textwidth}
  \centering
  \includegraphics[width=\linewidth, height=4.15cm]{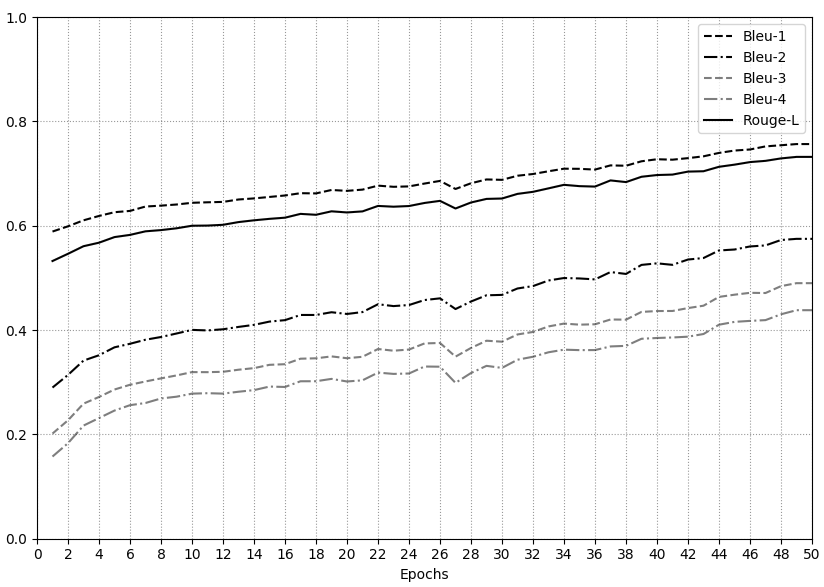}
  \caption{Reward during Training Process.}
  \label{fig:plots}
 \end{subfigure}
 \caption{}
\end{figure*}
  
\noindent
\textbf{Datasets:} The vocabulary of words obtained from the RSICD dataset contains a collection of 1653 words including the $<$start$>$, $<$unk$>$ and $<$end$>$ tokens. The dataset contains image-caption pairs of 30 classes namely airport, bare-land, baseball-field, beach, church, commercial, dense-residential, meadow, river, bridge etc. The dataset contains a total of 10921 images, split into 8734 training, 1094 validation, and 1093 testing images. The UCM-captions dataset contains identical captions for images from the same class and thus spans a vocabulary of only 210 words. It contains 21 classes land use image, including agricultural,
airplane, buildings, chaparral, overpass, parking lot, river, runway etc and with 100 images for each class. Because of the broad variety of image classes they provide, the above two datasets are widely used in remote sensing image captioning research.

\noindent
\textbf{Experiment Details:} We observed a substantial improvement in performance for extracting features from images by employing AlexNet \cite{NIPS2012_4824}. A fully connected layer with dimension of 256 and batch normalization along with the ReLU activation function is applied to the output of the feature extractor. We use an embedding module of dimension 256 to encode each word into an embedding space. 
For the value Network, a fully connected layer of dimension $ 256 \times 1$, with hyperbolic tan has been applied as the activation function to obtain a value function in the range [-1, 1]. 
We also observed that by employing a GRU in the encoder-decoder critic instead of an RNN results in a sharp increment in training speed. The functions $\psi_{1}$ and $\psi_{2}$ in Equation \ref{equations_enc} is a fully connected layer with dimension 256. 
We used the Adam Optimiser \cite{kingma2014method} with a learning rate of 5e-4 and is decreased by factor of 0.9 after every 10 iterations. The networks are trained for a total of 100 epochs.

\begin{figure*}[t!]
\begin{subfigure}{.2\textwidth}
\vspace{-18pt}
  \centering
  \includegraphics[width=0.8\linewidth]{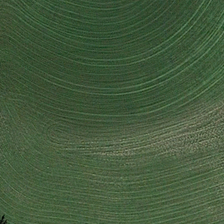}
  \caption { \textbf{A2C:} It is a piece of green meadow. \\
  \textbf{Ours:} A dirt lines are in this meadow.}
  \label{fig:meadow}
\end{subfigure}%
\hspace{30pt}
\begin{subfigure}{.2\textwidth}
\vspace{-13pt}
  \centering
  \includegraphics[width=0.8\linewidth]{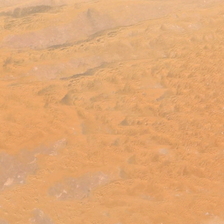}
  \caption{\textbf{A2C:} It is a piece of yellow desert.\\
  \textbf{Ours:} It is a rather flat desert stained with several black stains}
  \label{fig:desert}
\end{subfigure}
\hspace{30pt}
\begin{subfigure}{.2\textwidth}
  \centering
  \includegraphics[width=0.8\linewidth]{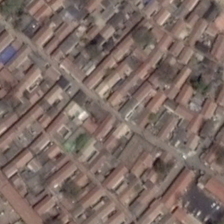}
  \caption{ \textbf{A2C:} Many rectangular buildings and green trees are in a dense area.
  \\ 
  \textbf{Ours:} Houses with red roofs on both sides of the road.
}
  \label{fig:residential}
\end{subfigure}
\hspace{30pt}
\begin{subfigure}{.2\textwidth}
\vspace{-18pt}
  \centering
  \includegraphics[width=0.8\linewidth]{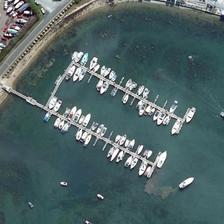}
  \caption{\textbf{A2C:} Many white boats are in the port. \\
  \textbf{Ours:} Two rows of white boats are in port}
  \label{fig:port}
\end{subfigure}%
\caption{Qualitative results of image captioning of the ADC setup on the RSICD dataset}
\label{fig:A2CvsADC}
\end{figure*}

\begin{figure*}[t!]
\begin{subfigure}{.2\textwidth}
  \centering
  \includegraphics[width=0.8\linewidth]{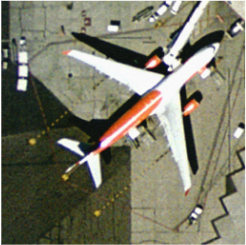}
  \caption { \textbf{A2C:} There are many airports at the airport.  \\
  \textbf{Ours:} There is a red airplane with lots of cars. }
  \label{fig:meadow}
\end{subfigure}%
\hspace{30pt}
\begin{subfigure}{.2\textwidth}
  \centering
  \includegraphics[width=0.8\linewidth]{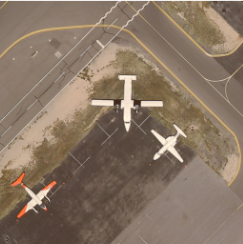}
  \caption{\textbf{A2C:}There are many airports at the airport. \\
  \textbf{Ours:} There is a red airplane in the airport. }
  \label{fig:desert}
\end{subfigure}
\hspace{30pt}
\begin{subfigure}{.2\textwidth}
  \centering
  \includegraphics[width=0.8\linewidth]{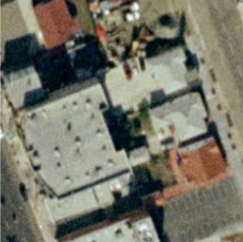}
  \caption{ \textbf{A2C:} There are many buildings. \\ 
  \textbf{Ours:} There is one road next to many buildings.
}
  \label{fig:residential}
\end{subfigure}
\hspace{30pt}
\begin{subfigure}{.2\textwidth}
  \centering
  \includegraphics[width=0.8\linewidth]{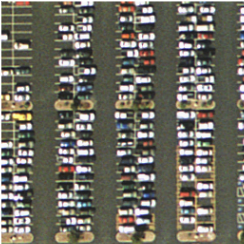}
  \caption{\textbf{A2C:} There are lots of cars with some buildings. \\
  \textbf{Ours:} Lots of cars are rectangular and close to each other in the parking lot.}
  \label{fig:port}
\end{subfigure}%
\caption{Qualitative results of image captioning of the ADC setup on the UCM-captions dataset}
\label{fig:A2CvsADConUCM}
\end{figure*}

\noindent
\textbf{Metrics for Comparison of results:} We compare our results qualitatively and quantitatively with the \cite{lu2017exploring} who trained a deep multi-modal neural network (referred to as MM) with different types of CNNs, RNNs, and LSTMs for semantic understanding of high-resolution remote sensing images in the RSICD and UCM-captions dataset. The authors also performed experiments on their dataset using the "hard" and "soft" attention mechanisms proposed by \cite{10.5555/3045118.3045336} denoted by HA and SA respectively. These three methods are not methods based on reinforcement learning, except for the mechanism of hard attention which uses the REINFORCE algorithm for attention but not for the prediction of the caption. Table \ref{Tab:table_res} and Table \ref{Tab:tableUCM} quantitatively compares the three methods (denoted by MM, SA, HA respectively) and the results of the A2C training setup with our proposed method. In Figure \ref{fig:A2CvsADC} and Figure \ref{fig:A2CvsADConUCM}, we compare the results of the A2C training setup replacing their LSTM by a Layer Norm LSTM like in our ADC training setup. Figure \ref{fig:plots} shows the values of the reward metrics ROUGE-L, BLEU-1, BLEU-2, BLEU-3 and BLEU-4 during training on the RSICD dataset. 

\begin{figure*}[t!] 
  \centering
  \includegraphics[width=\linewidth, height=8cm]{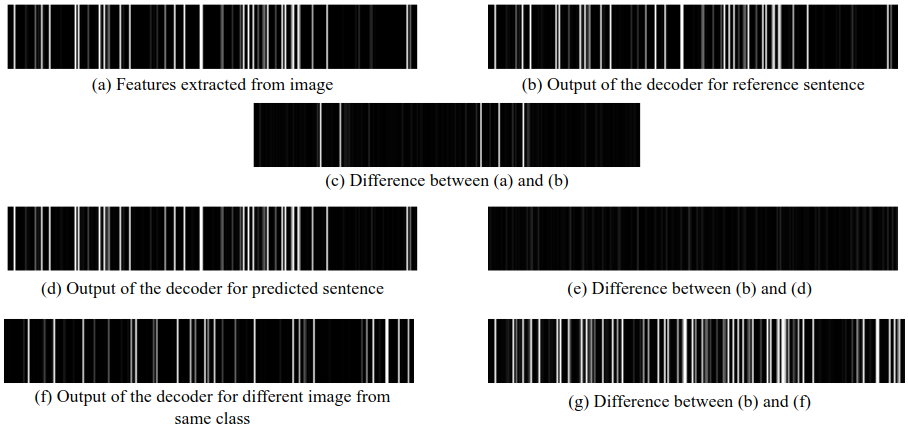}
  \caption{Qualitative results of the experiment comparing output of the decoder for different image-sentence pairs}
  \label{fig:qual}
\end{figure*}

\noindent
\textbf{Demonstrating validity of proposed critic:} The aim of this experiment is to visualise the relevance of the output of the decoder with respect to the output of the feature extractor CNN for different image-sentence pairs. To verify if the critic can distinguish between correct and incorrect captions, we also input a test image of the same class with a different sentence from the reference sentence of the ground truth (Figure \ref{fig:example_image}).
The figures in this section are normalized vector representation resized for the sake of visualization. Figure \ref{fig:qual}c represents the difference between the output of the decoder for a reference sentence and the features extracted by the CNN. The difference between the output of the decoder for the ground truth and predicted sentence is shown in Figure \ref{fig:qual}e. As expected, this difference is negligible (faint grey lines). This means that a generalized high-level correspondence between images and sentences that defines it semantically has been learned by this critic. Figure \ref{fig:qual}g is the difference between the representations learnt by the encoder-decoder critic for an image from the same class on a different ground truth caption. As expected, this difference is high implying that a sentence not describing an image does not get translated into abstract features corresponding to that image. We deduce that this critic does not trivially learn the identity function with respect to input features, due to the presence of non-invertible functions $\psi_{1}$ and $\psi_{2}$ in Equation \ref{equations_enc}. It means that the critic takes into account the correspondences of both image and sentence for two sentences with entirely different word2vec representations. We may infer from the above experiment that the critic effectively learns the correlations between images and reference sentences and encourages the generation of different sentences that encapsulate the same correlations.


\begin{figure*}[t!]
\begin{subfigure}{.2\textwidth}
  \centering
  \includegraphics[width=0.8\linewidth]{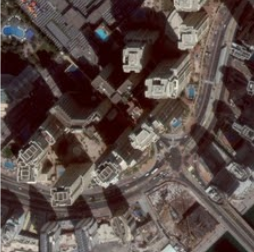}
  \caption { \textbf{Original:} Many tall buildings are in a commercial area. \\
  \textbf{Ours:}There is one road next to many buildings.}
  \label{fig:meadow}
\end{subfigure}%
\hspace{30pt}
\begin{subfigure}{.2\textwidth}
  \centering
  \includegraphics[width=0.8\linewidth]{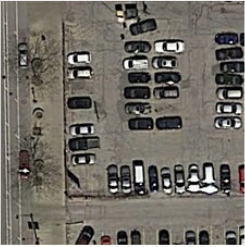}
  \caption{\textbf{Original:} Many cars are parked in the parking lot. \\
  \textbf{Ours:} Lots of cars are parked neatly in a parking lot. }
  \label{fig:desert}
\end{subfigure}
\hspace{30pt}
\begin{subfigure}{.2\textwidth}
  \centering
  \includegraphics[width=0.8\linewidth]{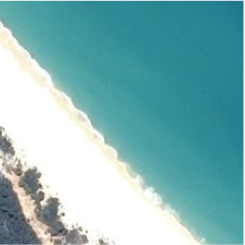}
  \caption{ \textbf{Original:} Yellow beach is between green ocean and green trees.\\ 
  \textbf{Ours:} This is a beach with blue sea and white sands. 
}
  \label{fig:residential}
\end{subfigure}
\hspace{30pt}
\begin{subfigure}{.2\textwidth}
  \centering
  \includegraphics[width=0.8\linewidth]{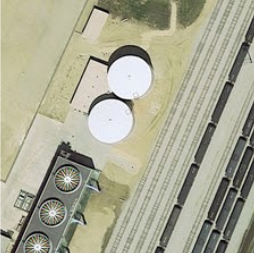}
  \caption{\textbf{Original:}On the ground, there are two spherical storage tank. \\
  \textbf{Ours:} Two small storage tanks are on the ground.
}
  \label{fig:port}
\end{subfigure}%
\caption{Qualitative results of cross dataset image captioning of the ADC setup on the RSICD dataset }
\label{fig:A2CvsADCcross}
\end{figure*}
\begin{figure*}
\begin{subfigure}{.2\textwidth}
\vspace{-13pt}
  \centering
  \includegraphics[width=4.8\linewidth,height=5.0cm]{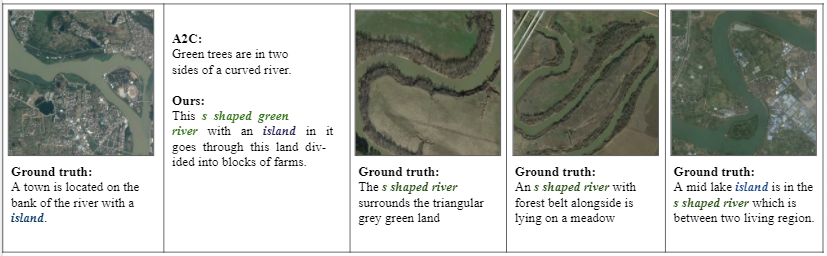}
  \label{fig:river}
\end{subfigure}
\\
\begin{subfigure}{.2\textwidth}
  \centering
  \includegraphics[width=4.8\linewidth,height=5.0cm]{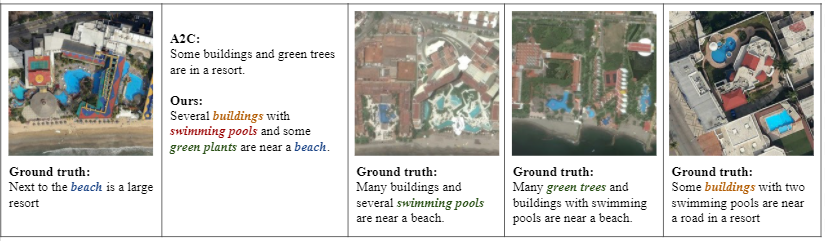}
  \label{fig:resort}
\end{subfigure}
\\
\begin{subfigure}{.2\textwidth}
  \centering
  \includegraphics[width=4.8\linewidth,height=5.0cm]{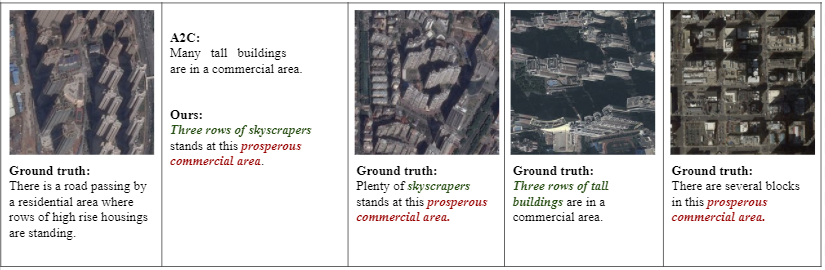}
  \label{fig:comm}
\end{subfigure}%
\\
\begin{subfigure}{.2\textwidth}
  \centering
  \includegraphics[width=4.8\linewidth, height=5.5cm]{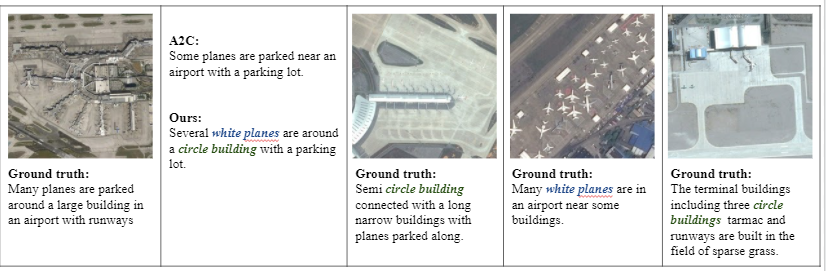}
  \label{fig:airport}
\end{subfigure}%
\caption{Qualitative comparision of results of image captioning of the ADC setup on the RSICD dataset. }
\label{fig:DetailedAnalysis}
\end{figure*}

\noindent
\textbf{Qualitative Analysis of Results:} As observed from the result of the sentence generation, our proposed method can generate more complicated yet explainable sentences having longer lengths, more
words with low frequency in caption labels. From the examples in Figure \ref{fig:DetailedAnalysis},  it is evident that the generated captions contain phrases that provide a highly accurate semantic explanation of the nature and localization of objects in the scene. We note, however, that these phrases are absent in the caption of ground truth sentences. We analyzed ground-truth captions of other images of the same class containing such phrases to explain the occurence of these phrases, and compared the test image with the corresponding images in the data set. From Figure \ref{fig:DetailedAnalysis}, it can be deduced that the addition of an encoder-decoder RNN critic has significantly improved the quality of the policy's performance on remote sensing images as compared to the baseline method (A2C). This demonstrates that the policy has successfully investigated the environment consisting of images and captions and has gained more knowledge compared to the baseline approach due to this critic's extra upgrade step in the optimization of policy objective. As observed in Table \ref{Tab:table_res}, our proposed approach provides a rapid increase in six out of seven scores used for comparison for the RSICD dataset. However, our approach proves superior for the UCM-captions dataset than previous methods for all the seven scores. CIDEr captures consensus-based human judgment better than established metrics through sentences created from different sources. The implementation of an encoder-decoder LSTM critic has resulted in strong increments in CIDEr for both the datasets.

\noindent
\textbf{Cross Dataset Captioning:} Testing trained models across datasets with similar domains gives an understanding of the model's ability to generalize and utilise it for real time predictions. We have therefore tested a model trained on the UCM-captions dataset on the RSICD dataset and made qualitative as well as quantitative comparisons with previous methods \cite{lu2017exploring}. The models trained on another dataset experience a rapid decline in metrics compared with the outcome of model trained on the corresponding dataset. From Table \ref{Tab:tableCross}, it is noted that the inclusion of our proposed critic has resulted in a substantial improvement in all 7 scores for the captioning of the cross datasets. We also note from Figure \ref{fig:A2CvsADCcross} that our policy generates the captions which convey the same meaning as the sentences of ground truth. This ensures that the policy trained using the ADC system can be used in Remote Sensing applications in real life.

\section{Conclusion}
In this paper, we proposed an  Actor Dual-critic (ADC) method for Image Captioning for the Remote Sensing Image Captioning Dataset. We are introducing another critic to the A2C training setup to encourage the prediction of sentences capturing relevant details along with sentence diversity. In the sense of converting sentences back to original images we suggest an encoder-decoder model for this critic. We also proposed a training strategy with an advantage factor based on a weighted difference of cosine similarities to update the policy parameters. We prove the superiority of our method quantitatively using the metrics BLEU-1, BLEU-2, BLEU-3, BLEU-4, ROUGE-L, and CIDEr. Ultimately, we use a model trained over the UCM-captions dataset and validate its superiority over other approaches to generate textual descriptions of images in the RSICD dataset.

\bibliographystyle{IEEEtran}
\bibliography{root}

\end{document}